%% file: main.tex
\documentclass[runningheads]{llncs}
\usepackage{graphicx}
\usepackage{tikz}
\usepackage{comment}
\usepackage{amsmath,amssymb} % define this before the line numbering.
\usepackage{color, colortbl}
\usepackage{arydshln}
\usepackage{subcaption}
\usepackage{wrapfig,booktabs,lipsum}
\definecolor{LightGray}{rgb}{0.92,0.92,0.92}
\definecolor{Red}{rgb}{1.0, 0.13, 0.32}

\usepackage[accsupp]{axessibility}  % Improves PDF readability for those with disabilities.

\usepackage{multirow,booktabs}
\usepackage[normalem]{ulem}
\usepackage{footnote}
\usepackage{kotex}
\usepackage{array}
\usepackage{adjustbox}
\usepackage{caption}
\usepackage{pifont}% http://ctan.org/pkg/pifont
\newcommand{\cmark}{\ding{51}}%
\newcommand{\xmark}{\ding{55}}%
\usepackage{float}

\usepackage{enumitem}

\usepackage[pagebackref=true,breaklinks=true,letterpaper=true,colorlinks,bookmarks=false]{hyperref}
\usepackage[capitalise]{cleveref}                 % Best reference package
\include{math_commands}

\usepackage{algorithm}
\usepackage{algpseudocode}

\usepackage{listings}
\definecolor{codegreen}{rgb}{0,0.6,0}
\definecolor{codegray}{rgb}{0.5,0.5,0.5}
\definecolor{codepurple}{rgb}{0.58,0,0.82}
\definecolor{backcolour}{rgb}{0.95,0.95,0.92}
\lstdefinestyle{mystyle}{
    backgroundcolor=\color{backcolour},   
    commentstyle=\color{codegreen},
    keywordstyle=\color{magenta},
    numberstyle=\tiny\color{codegray},
    stringstyle=\color{codepurple},
    basicstyle=\ttfamily\footnotesize,
    breakatwhitespace=false,         
    breaklines=true,                 
    captionpos=b,                    
    keepspaces=true,                 
    numbers=left,                    
    numbersep=5pt,                  
    showspaces=false,                
    showstringspaces=false,
    showtabs=false,                  
    tabsize=2
}
\begin{document}
% \linenumbers
\pagestyle{headings}
\mainmatter
\def\ECCVSubNumber{509}  % Insert your submission number here

\title{Multimodal Semi-Supervised Learning for Text Recognition} % Replace with your title

\begin{comment}
\titlerunning{ECCV-22 submission ID \ECCVSubNumber} 
\authorrunning{ECCV-22 submission ID \ECCVSubNumber} 
\author{Anonymous ECCV submission}
\institute{Paper ID \ECCVSubNumber}
\end{comment}
%******************

% \begin{comment}
\titlerunning{Multimodal Semi-Supervised Learning for Text Recognition}
% \author{Aviad Aberdam\\
% AWS AI\\
% {\tt\footnotesize aaberdam@amazon.com}
% \and
% Roy Ganz\\
% Technion\\
% {\tt\footnotesize ganz@cs.technion.ac.il}
% \and
% Shai Mazor\\
% AWS AI\\
% {\tt\footnotesize smazor@amazon.com}
% \and
% Ron Litman\\
% AWS AI\\
% {\tt\footnotesize litmanr@amazon.com}
% }

\author{
Aviad Aberdam\inst{1} \and
Roy Ganz\inst{2}\thanks{Work done during an Amazon internship.} \and
Shai Mazor\inst{1} \and
Ron Litman\inst{1}
}
\authorrunning{A. Aberdam et al.}
\institute{
AWS AI Labs \and Technion, Israel \\
\email{\{aaberdam,smazor,litmanr\}@amazon.com}, ~
\email{ganz@cs.technion.ac.il}}

% \end{comment}
%******************
\maketitle
\newcommand{\AlgoName}{SemiMTR}

\begin{abstract}
Until recently, the number of public real-world text images was insufficient for training scene text recognizers. Therefore, most modern training methods rely on synthetic data and operate in a fully supervised manner. Nevertheless, the amount of public real-world text images has increased significantly lately, including a great deal of unlabeled data. Leveraging these resources requires semi-supervised approaches; however, the few existing methods do not account for vision-language multimodality structure and therefore suboptimal for state-of-the-art multimodal architectures. To bridge this gap, we present semi-supervised learning for multimodal text recognizers (SemiMTR) that leverages unlabeled data at each modality training phase. Notably, our method refrains from extra training stages and maintains the current three-stage multimodal training procedure. Our algorithm starts by pretraining the vision model through a single-stage training that unifies self-supervised learning with supervised training. More specifically, we extend an existing visual representation learning algorithm and propose the first contrastive-based method for scene text recognition. After pretraining the language model on a text corpus, we fine-tune the entire network via a sequential, character-level, consistency regularization between weakly and strongly augmented views of text images. In a novel setup, consistency is enforced on each modality separately. Extensive experiments validate that our method outperforms the current training schemes and achieves state-of-the-art results on multiple scene text recognition benchmarks. Code will be published upon publication.

\end{abstract}

\section{Introduction}

Understanding written information through vision is essential to most of our everyday tasks, and as such, it is a key research area in artificial intelligence. For that reason, scene text recognition, which deals with text in natural environments, has been studied extensively in the past two decades \cite{Long2018survey,chen2020text,sengupta2020journey}.
Unfortunately, until recently, public real-world data for scene text have been fairly scarce, leading the community to focus almost entirely on synthetic training data and hence on supervised training methods~\cite{shi2016end,ASTER,SAR,ESIR,Mask-textspotter,Baek2019clova,ScRN,DAN,textscanner,srn_yu2020towards,litman2020scatter,qiao2020seed,robustscanner,mou2020plugnet}.
However, in recent years, the number of public real-world text images has increased significantly, including a few million unlabeled images.
Therefore, as advocated by~\cite{whatif_baek2021if}, it is time to focus on unsupervised and semi-supervised algorithms that can advantage these resources. 
Such methods, though, have been barely studied for text recognition~\cite{zhang2019sequence,seqclr_aberdam2021sequence,whatif_baek2021if,abinet_fang2021read}, while the few that do exist do not account for the multimodal structure and therefore cannot fully leverage unlabeled data in advanced multimodal recognizers~\cite{wang2021two,abinet_fang2021read,qiao2020seed,srn_yu2020towards}.

In this work, we present semi-supervised learning for multimodal text recognizers (\AlgoName{}). To this end, we adopt the ABINet~\cite{abinet_fang2021read} vision-language multimodal architecture and introduce a semi-supervised training scheme that utilizes unlabeled data at each modality training and therefore leverages the multimodal structure.
Originally, ABINet is trained via three stages: (i) a supervised vision model pretraining; (ii) a bidirectional language representation learning of the language model; and (iii) a supervised fine-tuning of the fusion model and entire network.
Our method offers different first and third steps for this procedure that can benefit from unlabeled real-world images. In contrast to leading semi-supervised text recognition approaches, which are based on pseudo-labeling and require additional retraining steps~\cite{whatif_baek2021if,abinet_fang2021read}, our training procedure maintains the original three-stage scheme.

For pretraining the vision model, we harness both labeled and unlabeled data via contrastive learning, which, as we show, leads not only to improved performance of the vision model but also of the entire end-goal network.
The key idea of contrastive-based methods is to maximize agreement between representations of differently augmented views of the same image and distinguish them from representations of other images~\cite{chen2020simple,he2020momentum}. This way, the visual backbone learns to concentrate on the image text content and ignore unimportant attributes, such as color, texture, and perspective.
More specifically, we adapt SeqCLR~\cite{seqclr_aberdam2021sequence}, an existing sequence-to-sequence contrastive learning method, originally designed for handwriting recognition, and extend it to scene text recognition.
For this goal, we adopt a more robust transformer-based backbone, apply stronger color-texture augmentations, which are essential to text-in-the-wild images, and present a single training stage that already contains the supervised training, which usually operates separately after.
Our vision-model pretraining is the first contrastive method successfully applied to scene text images.

Our fine-tuning stage, which targets the final model predictions, presents a sequential, character-level, consistency regularization between weakly and strongly augmented views of real-world text images. Our scheme first generates a sequence of artificial pseudo-labels from a weakly-augmented text image and then uses it to train all modalities fed by a strongly-augmented view of the same image. Our study reveals that this learning is most beneficial when each modality is the teacher of itself and generates its pseudo-labels to train on.

We first validate our method experimentally against state-of-the-art synthetic-based methods and reveal that applying our algorithm only on real-world images reduces the word error rate by 14\% on widely-used scene text benchmarks and 28\% on non-common ones. Our results are the first to surpass synthetic-based solutions and, thus, significantly strengthen the claim of \cite{whatif_baek2021if} that real datasets have been accumulated to a sufficient level.
In addition, we compare our scheme with existing semi-supervised methods and demonstrate improvements in the overall performance, enhancing the state-of-the-art on 12 out of 13 scene text recognition benchmarks.
Finally, we provide a comprehensive analysis of our framework, exploring the impact of its components. 

To summarize, the main contributions of our work are:
\begin{itemize}
    \item A multimodal semi-supervised learning algorithm for text recognition, which is customized for modern vision-language multimodal architectures.
    
    \item A unified one-stage pretraining method for the vision model, which is the first successful contrastive learning scheme for scene text recognition.
    
    \item A sequential, character-level, consistency regularization in which each modality teaches itself.
    
    \item Extensive experiments demonstrate state-of-the-art performance on multiple scene text recognition benchmarks.
\end{itemize}

\section{Related Work}
\paragraph{\textbf{Vision-language multimodal text recognizers.}}
In the past few years, modern text recognition schemes have focused on extracting visual and semantic information encompassed in text images and balancing them together~\cite{qiao2020seed,srn_yu2020towards,litman2020scatter,abinet_fang2021read,wang2021two}.
More specifically, SCATTER~\cite{litman2020scatter} proposed custom decoders for these two information types; SRN~\cite{srn_yu2020towards} introduced a global semantic reasoning module; SEED~\cite{qiao2020seed} offered a semantics
enhanced encoder-decoder framework; and VisionLAN~\cite{wang2021two} endued the vision
model with language capability. 
% All three methods generate the final network prediction by fusing the visual and lingual information.
Recently, ABINet~\cite{abinet_fang2021read} presented a vision-language multimodal architecture that possesses an explicit language model, which can be pretrained on some text corpus. Through a three-stage training procedure described in \cref{sec:text_recog_background}, ABINet reached state-of-the-art results on scene text recognition.
Our work is the first to present a semi-supervised learning scheme that utilizes the vision-language structure in these modern text recognition methods.

\paragraph{\textbf{Semi-supervised learning for text recognition.}}
Untill recently, real-world public images of scene text were so rare, that the few to exist were used for evaluation. Therefore, most training methods~\cite{shi2016end,ASTER,SAR,ESIR,Mask-textspotter,Baek2019clova,ScRN,DAN,textscanner,srn_yu2020towards,litman2020scatter,qiao2020seed,robustscanner,nuriel2021textadain} have been based on two large synthetic datasets~\cite{jaderberg2014synthetic,gupta2016synthetic} and have operated in a fully-supervised fashion.
Exceptions to this are \cite{zhang2019sequence} and~\cite{kang2020unsupervised}, which suggested sequence-to-sequence domain adaptation techniques between labeled and unlabeled datasets, and \cite{seqclr_aberdam2021sequence} which introduced a sequence-to-sequence contrastive learning for visual representations learning.
In addition, \cite{whatif_baek2021if,janouskova2021text,abinet_fang2021read} offered pseudo-labeling methods to utilize real-world unlabeled images, while the latter added a confidence-based criterion for filtering noisy pseudo-labels.
Our work is the first to propose semi-supervised learning for text recognition that utilizes the multimodal structure. 
% To this end, we develop self-supervised learning methods for the vision model pretraining, which is based on representation learning, and for the fine-tuning stage, which is based on consistency regularization.
To this end, we develop self-supervised learning methods for the vision model pretraining and the fine-tuning phase, based on representation learning and consistency regularization, respectively.

\paragraph{\textbf{Consistency regularization.}}
Consistency regularization is a widely-used technique in self-supervised learning. Its core idea is that the model predictions should remain the same under each semantic-preserving perturbations of the same image. Many works \cite{bachman2014learning,sajjadi2016regularization,laine2016temporal,tarvainen2017mean,berthelot2019mixmatch,xie2020unsupervised} introduce mechanisms to impose consensus on the network outputs.
For example, \cite{sajjadi2016regularization} enforces the network to be agnostic to some transformations and disturbances, and~\cite{xie2020unsupervised} proposes a consistency regularization in semi-supervised settings by using noise injections and augmentations on the unlabeled examples. 
A more closely-related work to our proposed method is FixMatch~\cite{sohn2020fixmatch}. This paper integrates consistency regularization with pseudo-labeling to benefit from unlabeled data in classification tasks. More specifically, FixMatch trains the predictions from the strongly-augmented version to match the pseudo-label produced from the weakly-augmented view of the same image. 
Motivated by these papers, we propose a sequential, character-level, consistency regularization for the fine-tuning stage of the whole network, in which we enforce consistency on each modality separately. Our scheme, tailored for the modern multimodal architectures, outperforms current semi-supervised methods for scene text recognition.

\section{Architectural Background}
\label{sec:text_recog_background}

\begin{figure}[t]
\normalsize
  \centering
  \includegraphics[width=0.95\textwidth]{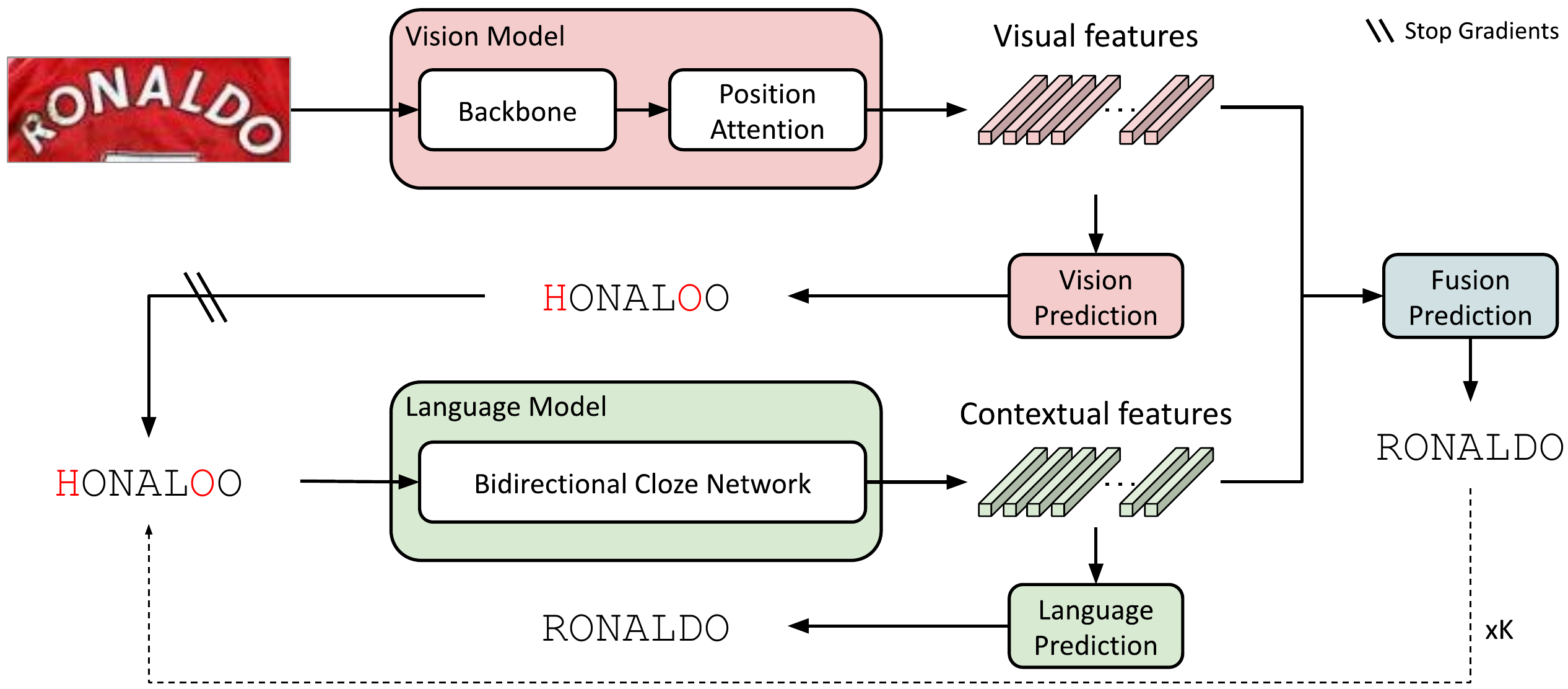}
  \caption{\textbf{Text recognition architecture}. We adopt ABINet~\cite{abinet_fang2021read} as our case study of vision-language multimodal recognizers.
  In this scheme, the vision model first extracts a sequence of visual features from a given image and then decodes it into character predictions. Next, these predictions are fed to the language model, which derives contextual features. Finally, the fusion model operates on the visual and contextual features and provides the network final prediction. To further improve accuracy, an optional iterative phase, drawn with a dashed line, reinserts the output back to the language model as a predictions-refinement step}
  \label{fig:abinet_background}
  \vspace{-0.2cm}
\end{figure}

Throughout this work, we focus on the novel multimodal ABINet~\cite{abinet_fang2021read} architecture, on which we demonstrate our multimodal semi-supervised (\AlgoName{}) algorithm. As depicted in \cref{fig:abinet_background} and further detailed in \cref{app:abinet}, ABINet architecture fuses the features of the vision model and the language model to obtain its final predictions. ABINet's training scheme comprises three stages. The first stage is a fully-supervised pretraining of the vision model. Independently, we pretrain the language model on a large text corpus via a version of masked language modeling~\cite{devlin2018bert}, in which we mask the attention maps instead of input characters. Finally, an end-to-end fine-tuning stage is applied via supervised cross-entropy losses on the vision, language, and fusion predictions.
The following section describes our training method, which enables leveraging unlabeled text images in multimodal architectures.

\section{\AlgoName: Multimodal Semi-Supervised Learning}
\label{sec:mumss_framework}

We introduce \AlgoName{}, semi-supervised learning for multimodal text recognizers, which aims to fully utilize vision-language multimodal architectures by leveraging unlabeled data at each modality training. Our method offers an overall simple algorithm that maintains a three-stage training procedure as in ABINet supervised learning (see \cref{sec:text_recog_background} above). Our vision model pretraining is an extended version of the sequence-to-sequence contrastive learning~\cite{seqclr_aberdam2021sequence}, which we integrated with a supervised loss. 
Independently, the language modality is trained on a large text corpus using the same masked language modeling of ABINet~\cite{abinet_fang2021read}. 
For completeness, we further describe this step in \cref{app:language_pretraining}.
Finally, the entire network is fine-tuned via a sequential, character-level, consistency regularization between weakly and strongly augmented views of the input images. We empirically find out that this regularization is the most effective when each modality is the teacher of itself and creates its own pseudo-labels.

\subsection{Vision Model Pretraining}
\label{sec:vision_pretaining}
To better utilize unlabeled real data, we propose including these resources also during the vision model pretraining.
As our experiments show, such training not only improves the accuracy of the vision model but eventually leads to better generalization of the final network and performance gain over competitive semi-supervised methods~\cite{abinet_fang2021read,whatif_baek2021if}, which use unlabeled data only in the last training phase.
These results indicate that the semi-supervised learning of the vision model is essential to fully benefit from unlabeled data in multimodal schemes.

\begin{figure}[t]
\normalsize
  \centering
  \includegraphics[width=0.9\textwidth]{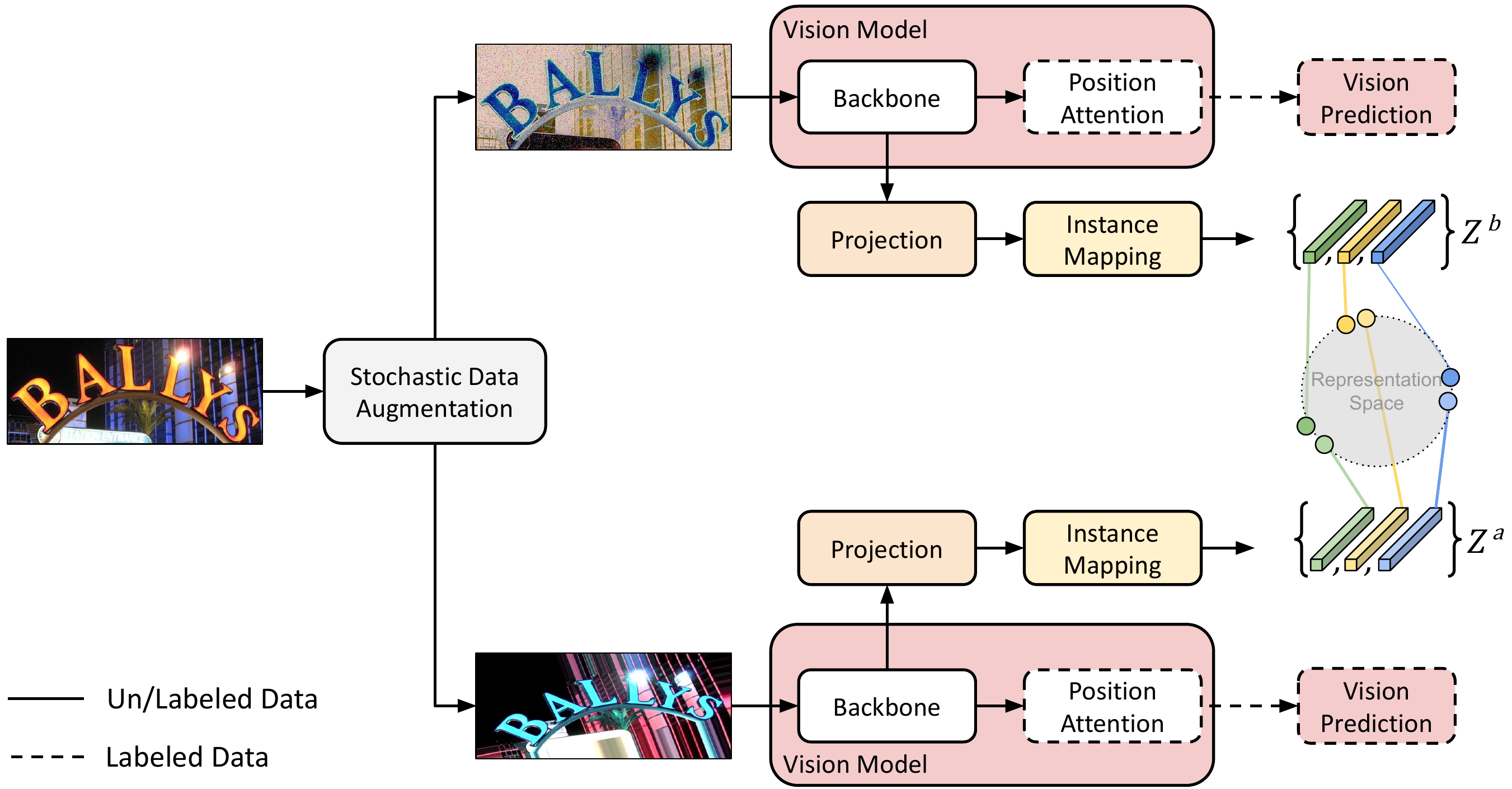}
  \caption{\textbf{Visual representation learning for vision model pretraining.} Our method augments twice each image in a batch and feeds these views into a visual backbone and projection head. Next, we apply an instance-mapping function~\cite{seqclr_aberdam2021sequence}, which creates a sequence of representations for each augmented view and thus enables contrastive learning at a sub-word level. A parallel branch computes supervised loss on the visual predictions of the labeled data}
  \label{fig:main_fig_vision}
  \vspace{-0.2cm}
\end{figure}

As illustrated in \cref{fig:main_fig_vision}, this stage harnesses unlabeled data by performing unsupervised visual representation learning along with supervised training. In particular, we extend the sequence-to-sequence contrastive learning framework (SeqCLR)~\cite{seqclr_aberdam2021sequence} to scene text recognition and transformer-based backbone and unify it with supervised training.
Interestingly, as experimented in \cref{subsec:vision_pretraining_ablation}, integrating the visual representation learning with the supervised training into a single stage leads to performance gains.
More specifically, our pretraining procedure consists of the following six building blocks:
\begin{itemize}
    \item \textbf{Stochastic data augmentation}, which creates two augmented views for each input text image.
    The original augmentation pipeline proposed in SeqCLR~\cite{seqclr_aberdam2021sequence} was mainly designed for handwritten-text images, in which the background is usually solid and of light colors. However, in scene text images, the background significantly varies between different images. Therefore, to obtain representations that focus on the foreground text and not the background, we propose stronger augmentations, especially in terms of the image color and texture. Note that the proposed augmentations preserve the sequential structure of the input to maintain alignment between the two sequences of representations that will be extracted from these augmented views. \cref{fig:augmentation} in the ``Strong'' column presents augmentation pipeline demonstrations, and \cref{app:augmentations} provides its pseudo-code.
    
    \begin{figure}[t]
    \normalsize
      \centering
      \includegraphics[width=0.6\textwidth]{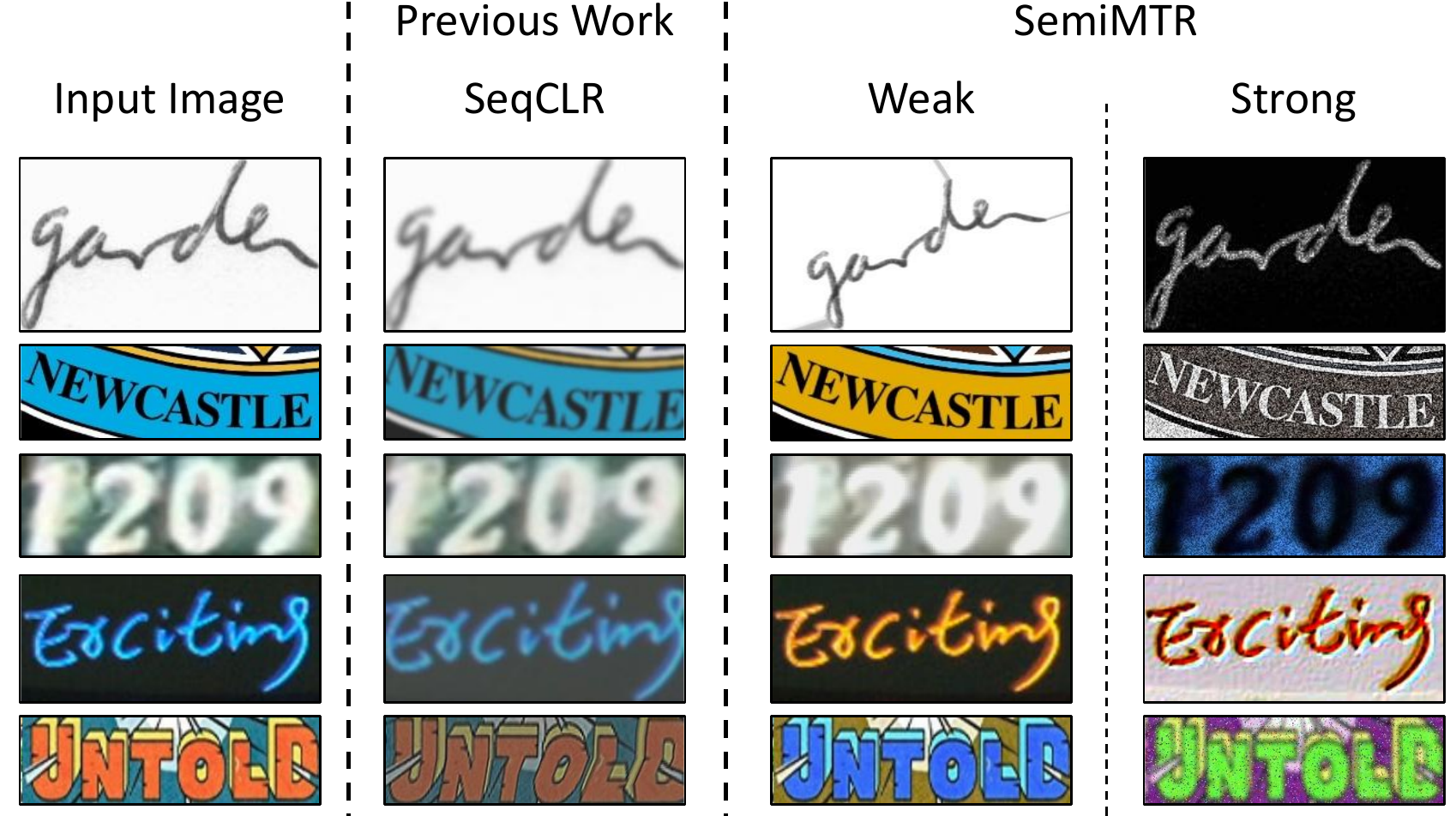}
      \caption{\textbf{Proposed augmentations.} The augmentation pipeline proposed by SeqCLR~\cite{seqclr_aberdam2021sequence} is mainly designed for handwriting text images, in which the background is often solid and of light colors. See an example in the first row. However, the background of scene text images varies significantly between different instances, as illustrated in the last four rows. Therefore, to enforce the learned representation to concentrate on the text content and not the image visual attributes, we propose higher severity augmentations, especially in terms of image color and texture
    %   . However, contrary to the supervised augmentations, in the self-supervised pipelines, we still maintain sequence-level misalignment as was proposed by \cite{seqclr_aberdam2021sequence}
        }
      \label{fig:augmentation}
    \end{figure}

    \item \textbf{Visual backbone} extracts visual features out of the augmented images. Unlike SeqCLR~\cite{seqclr_aberdam2021sequence}, which applied representation learning merely to a convolutional neural network (CNN), here, we follow the leading text recognition methods \cite{abinet_fang2021read,lee2020recognizing,atienza2021vision} and adopt the visual backbone of ABINet \cite{abinet_fang2021read}, which consists of ResNet and Transformer units.

    \item \textbf{Projection head}, an optional auxiliary network, which transforms the visual backbone features into a lower-dimensional space.

    \item \textbf{Instance-mapping function}. A unique block for sequence-to-sequence predictions divides each feature map into a sequence of separate representations over which the contrastive loss is computed. This operation enables contrasting between individual elements of the sequence and not whole images, as in standard contrastive methods \cite{chen2020simple,he2020momentum,grill2020bootstrap,caron2020unsupervised}. We adopt the window-to-instance mapping approach \cite{seqclr_aberdam2021sequence}, in which a batch of (possibly projected) feature maps, $\tP \in \R^{N \times F \times H \times W}$, is first flattened to the shape of $\R^{N \times H \cdot W \times F}$. Then, using the adaptive average pooling, we extract $T$ separate representations out of each flattened feature map and collect them into a set $\gZ$:
    \begin{equation}
        \gZ = \operatorname{AdaptiveAvgPool2d}(\operatorname{Flatten}(\tP)).
    \end{equation}
    This function is called twice for each augmented view of the input, resulting in two sets of representations, $\gZ^a, \gZ^b$.
    
    \item \textbf{Contrastive loss} aims to bring closer corresponding representations, termed positive pairs, and distinguish them from the others, denoted as negative examples. The contrastive loss is calculated over the two sets of representations extracted by the instance-mapping function, $\gZ^a, \gZ^b$, each of the size of $NT$:
    \begin{equation} \label{eq:seqclr_loss}
        \gL_{\text{SeqCLR}}(\gZ^a, \gZ^b) = \sum_{0 \leq i < NT} \ell_{\text{NCE}} \left( \rvz^a_i, \rvz^b_i; \, \gZ^a \cup \gZ^b \right) + \ell_{\text{NCE}} \left( \rvz^b_i, \rvz^a_i; \, \gZ^a \cup \gZ^b \right),
    \end{equation}
    where $\ell_{\text{NCE}}$ is the noise contrastive estimation (NCE) loss function~\cite{oord2018representation}:
    \begin{equation}
        \ell_{\text{NCE}}(\rvu^a, \rvu^b; \, \U) = -\log \frac{ \exp\left( \simop(\rvu^a, \rvu^b) / \tau \right)}{\sum_{\rvu \in \U\setminus \rvu^a}\exp\left(\simop(\rvu^a, \rvu) / \tau \right)},
    \end{equation}
    with a temperature parameter $\tau$ and similarity operator of the cosine distance, $\simop(\rvv, \rvu) = \rvv^T \rvu / \norm{\rvv} \norm{\rvu}$.
    
    \item \textbf{Vision decoder and supervised loss}. In SeqCLR~\cite{seqclr_aberdam2021sequence} and standard contrastive learning methods~\cite{chen2020simple,he2020momentum,grill2020bootstrap,caron2020unsupervised}, the supervised training phase occurs after the self-supervised stage ends. However, we refrain from increasing the overall training stages and offer an integrated, semi-supervised training scheme.
    Therefore, the overall loss of the vision model pretraining is:
    \begin{equation} \label{eq:vision_pretraining_loss}
        \gL = \lambda_U \sum_{\gD_U, \gD_L} \gL_{\text{SeqCLR}} + \lambda_L \sum_{\gD_L} \gL_{\text{ce}},
    \end{equation}
    where $\gD_U, \gD_L$ are the unlabeled and labeled datasets, $\gL_{\text{ce}}$ is the cross-entropy loss on the visual decoder predictions, and $\lambda_U, \lambda_L \geq 0$ are scalar coefficients.
    
\end{itemize}

% In parallel to the above pretraining, the language model is pretrained, as suggested by \cite{abinet_fang2021read}, on a text corpus via bidirectional representation learning. This step resembles the masked language model training; however, instead of masking each character at a time in the text instance, it masks the attention maps, which prevents each character prediction from seeing its own input. For completeness, we describe this step in \cref{app:language_pretraining}.

\subsection{Consistency Regularization for Fusion Model Training}
\label{sec:consistency_reg}

\begin{figure}[t]
\normalsize
  \centering
  \includegraphics[width=0.7\textwidth]{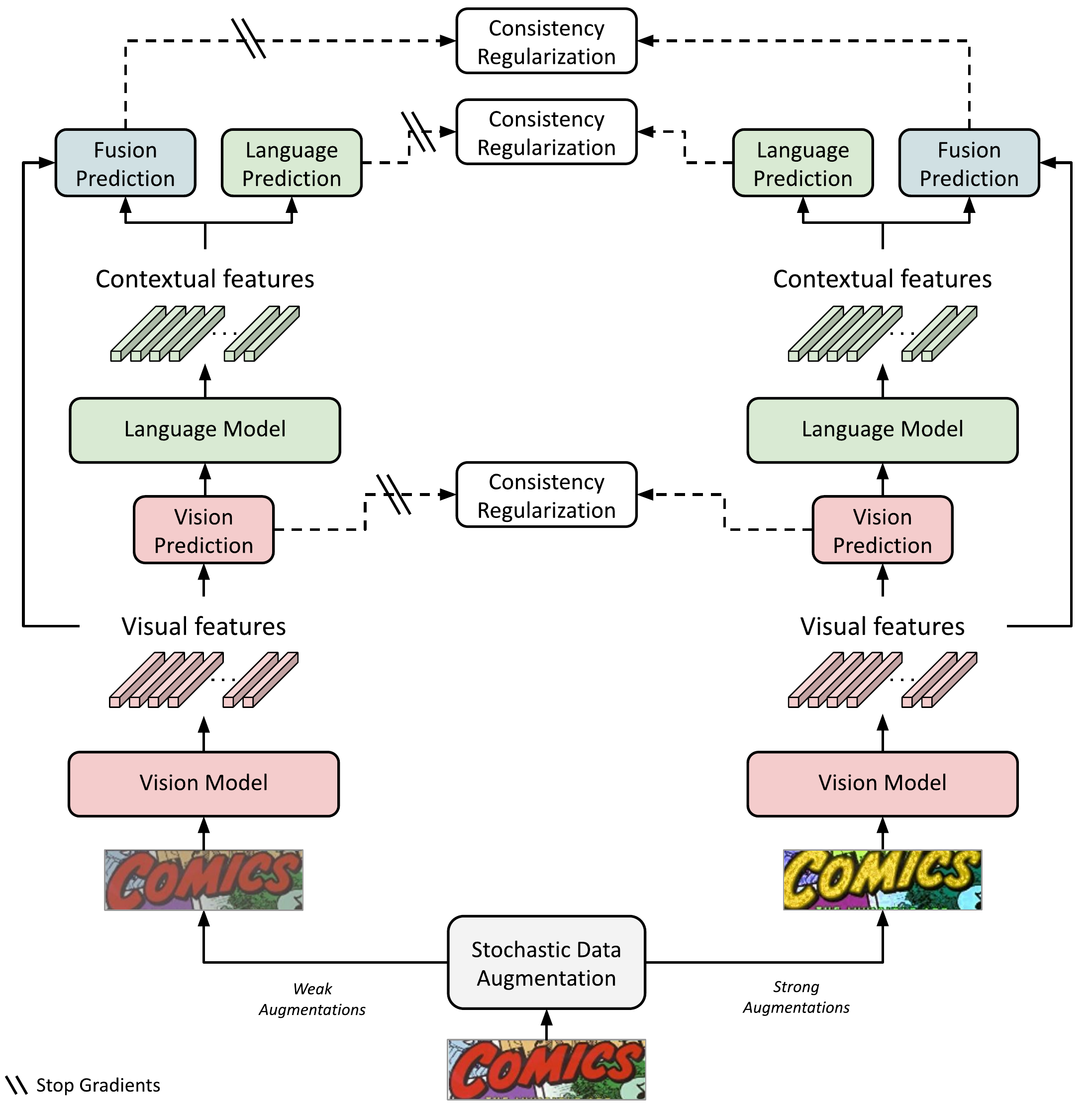}
  \caption{\textbf{Sequential consistency regularization for model fine-tuning.} Our scheme generates a sequence of artificial pseudo-labels for each modality out of a weakly-augmented image (left). Then, a sequential, character-level, consistency regularization is computed for each modality separately between the above pseudo-labels and the modality predictions for a strongly-augmented view of the same image (right). 
  We further calculate a supervised loss on the labeled data}
  \label{fig:consistency_reg}
\end{figure}

After pretraining the vision and language models, we move on to train the fusion model and fine-tune the entire network. Here, as well, we introduce a unified stage that combines both supervised and self-supervised training objectives. Therefore, this training offers a more efficient procedure than leading semi-supervised methods~\cite{abinet_fang2021read,whatif_baek2021if}, which are based on pseudo labeling and thus require retraining steps.
For the self-supervised learning at this stage, we propose a sequential, character-level, consistency regularization, which better suits the nature of the fine-tuning phase.
As demonstrated in \cref{fig:consistency_reg}, our method first generates a sequence of artificial labels out of a weakly-augmented version of the image and then uses them to train \emph{all modalities} fed by a strongly-augmented view of the same input.
Our work provides a comprehensive study of this stage, exploring each of the following components:
\begin{itemize}
    \item \textbf{Stochastic strong and weak augmentations}. We fix the weak augmentation pipeline to be as used in ABINet~\cite{abinet_fang2021read} for supervised training. For the strong augmentations, we examine different techniques and find out that the color-texture augmentation method, proposed for our vision model pretraining, is the most efficient.
    See samples of these augmentations in \cref{fig:augmentation}.
    
    \item \textbf{Sequential consistency regularization loss} is computed between sequential predictions of weakly and strongly augmented views of the same input image. Our method first prunes the teacher's sequential prediction at the first padding token location, denoted by $N^{\text{weak}}$. Then, it applies the consistency regularization on each character independently, using the configurable loss function of:
    \begin{equation}\label{eq:sequential_consistency}
        \gL_{\text{Consist}}(\rmY^{\text{strong}}; \rmY^{\text{weak}}) = \sum_{0 \leq i < N^{\text{weak}}} \1(\max(\rvy_i^{\text{weak}}) > t) \ell(\rvy_i^{\text{strong}}, \rvy_i^{\text{weak}}),
    \end{equation}
    where $\rmY^{\text{strong}}, \rmY^{\text{weak}}$ denote the sequential probability vectors of strongly and weakly augmented views, $\1(\cdot > t)$ is the threshold operator on $t \geq 0$, and $\ell$ is the core loss function, which in our case is cross-entropy or KL-divergence.
    In \cref{subsec:fusion_pretraining_ablation}, we examine the effect of these ingredients, including the core loss function, the threshold, and the type of the teacher labels, $\rmY^{\text{weak}}$ -- raw probability vectors versus one-hot labels.
    
    \item \textbf{Teacher and student modalities}. Usually, the teacher and student decoders on which the consistency regularization is computed are simply the same single decoder of the entire network. However, in multimodal text recognition schemes, each modality has its own decoder, and thus, can serve as either a teacher or a student. In \cref{subsec:fusion_pretraining_ablation}, we further examine these different configurations and find out that each modality should generate its own pseudo-labels, namely, be its own teacher. Therefore, the loss can be formulated as follows:
    \begin{equation}
    % \begin{split}
    \label{eq:overall_consistency_loss}
    \hspace*{-5mm}
        \sum_{d\in \text{decoders}}
        \lambda_{U_d} \sum_{\gD_U,\gD_L} \gL_{\text{Consist}}(\rmY_d^{\text{strong}}; \rmY_d^{\text{weak}}) + \lambda_{L_d} \sum_{\gD_L} \gL_{\text{ce}}(\rmY_d^{\text{strong}}) + \gL_{\text{ce}}(\rmY_d^{\text{weak}}),
    % \end{split}
    \end{equation}
    where $\text{decoders} = \{\text{vision},~\text{language},~\text{fusion}\}$, $\gD_U,\gD_L$ are unlabeled and labeled datasets, $\rmY_d^{\text{strong}}, \rmY_d^{\text{weak}}$ denote the probability maps of decoder $d$ from strongly and weakly augmented views, $\gL_{\text{ce}}(\cdot)$ is the supervised cross-entropy loss with the ground-truth labels, and $\lambda_{U_d}, \lambda_{L_d} \geq 0$ are coefficient scalars.
\end{itemize}

\section{Experiments}

In this section, we demonstrate the effectiveness of our method through extensive experiments, while our goal is two-fold. First, we aim to analyze the capability of current synthetic and real-world datasets. We show that applying our algorithm on real-world data surpasses all supervised synthetic-based methods, although real-world labeled data are only 1.7\% of synthetic data. 
Our work is the first to achieve such results, whereas \cite{whatif_baek2021if} only reached near but below state-of-the-art results. 
Specifically, our method reduces the word error rate by 14\% on standard benchmarks and 28\% on less common ones compared to ABINet~\cite{abinet_fang2021read}.
That said, synthetic data is still valuable, and when harnessed with real-world data, it further reduces the word error rate by 12\%.
In addition, we compare our method with leading semi-supervised methods, which, for a fair comparison, are reimplemented and applied on the same training set. \AlgoName{} consistently outperforms these methods on multiple scene text benchmarks.

\begin{wrapfigure}{R}{5cm}
\vspace{-0.8cm}
    \input{tables/dataset_info_sidebyside}
\vspace{-0.7cm}
\end{wrapfigure}

\paragraph{\textbf{Datasets.}} We examine our method on multiple public datasets of scene text recognition, which we list in \cref{tab:dataset_info} and detail in \cref{app:dtatsets}.
We denote by \emph{Real-L} and \emph{Real-U} the real-world labeled and unlabeled training datasets, as defined by \cite{whatif_baek2021if}, and by \emph{Synth} the synthetic datasets.
Unlike several works which used the test set for validation, our validation set is a portion of the training partitions of Real-L dataset, as divided by~\cite{whatif_baek2021if}.
In addition to the common six scene text benchmarks, IIIT~\cite{Mishra2012sj}, SVT~\cite{Wang2011bottom}, IC03~\cite{Lucas2003ic03} IC13~\cite{Karatzas2013ic13}, IC15~\cite{Karatzas2015ic15}, SVTP~\cite{Phan2013svtp} and CUTE~\cite{Risnumawan2014cute}, which we term \emph{Common Benchmarks}, we also consider the test partitions of Real-L, which we term \emph{Non-Common Benchmarks}.
Our report provides word-level accuracy on each of these datasets and weighted averages on the common and non-common benchmarks.
As a general statement, we believe that \emph{as a community, we should expand the scope of the evaluation benchmarks}. The reason is that common benchmarks are pretty small, only 7.6K images overall, and therefore are insufficient and may lead to false scientific discoveries.
The extreme case is the CUTE dataset, 288 images, on which the accuracy is already higher than 90\%, and therefore, new algorithms are measured on the misclassified examples, less than 30 images. 
On the other hand, adopting the non-common benchmarks significantly enlarges the test set and enriches its domain diversity.

\noindent \emph{\textbf{Implementation details.}}~
In many configuration parameters, such as architecture, optimizer, and image size, we follow the setting of ABINet~\cite{abinet_fang2021read}.
A full description is provided in \cref{app:implementation_details}.

\noindent \emph{\textbf{Training time.}}~ A comparison with ABINet supervised training~\cite{abinet_fang2021read} shows that our vision model pretraining is longer by 65\% (495K vs.~300K iterations), but our fusion model training is shorter by 65\% (130K vs.~376K). Note that the latter training involves the whole network, and thus each iteration takes longer.

% \subsection{Scene Text Recognition}
\input{tables/sota_all_scene_text_datasets_v2}

\subsection{Comparison to State-of-the-Art Methods}

Our study aims to probe the efficiency of current synthetic and real-world public datasets.
Therefore, we start by comparing training on synthetic images to real-world labeled data.
While synthetic data are relatively easy to acquire and lead to on-par performance on common benchmarks, they result in poor performance on the non-common benchmarks. See TRBA and ABINet results on Synth versus Real-L in \cref{tab:eval-str-sota}.
More specifically, ABINet trained on synthetic datasets achieves 52.0\% on non-common benchmarks, compared to 62.4\% of ABINet trained on Real-L.
This observation, also reported by \cite{whatif_baek2021if}, demonstrates the potential of synthetic data, but simultaneously indicates that current public, synthetic data do not represent well the diversity of the non-common benchmarks.

Building upon this understanding, we move on to experiment methods that can leverage the unlabeled real-world datasets (Real-U).
To this end, we compare our algorithm with the two recent semi-supervised approaches proposed in \cite{whatif_baek2021if,abinet_fang2021read}, which are both based on pseudo-labeling. Namely, a fully-supervised trained model produces pseudo-labels for the unlabeled data, and then the model is retrained on the labeled and pseudo-labeled data.
The difference between both methods is that \cite{abinet_fang2021read}, denoted by ABINet$_{est}$, filters out pseudo-labels of low confidence level, while \cite{whatif_baek2021if}, denoted as ABINet$_{PL}$, uses all the pseudo-labels.
For a fair comparison, we reimplement both training schemes and apply them to the same text recognition architecture and training sets.
Additional implementation details are listed in Appendix \ref{app:comparison_sota}.

As shown in \cref{tab:eval-str-sota}, leveraging the unlabeled real-world data significantly boosts performance, although these images mainly come (88\%) from book covers, a different domain to the test sets.
The comparison between the semi-supervised methods (Real-L + Real-U) shows that our method consistently improves performance across all the common and non-common benchmarks. This performance gain is achieved even though \AlgoName{} requires fewer training stages, as it does not involve a retraining stage.

We further analyze our method and examine the performance gain of leveraging unlabeled data only at the vision model pretraining, which we named \emph{SemiMTR-V}, and only at the fine-tuning stage, denoted as \emph{SemiMTR-F}. As presented in \cref{tab:eval-str-sota}, each of these parts efficiently utilizes unlabeled data. In fact, each of these partial versions of our algorithm reaches a higher overall average accuracy, calculated across common and non-common benchmarks, than the current semi-supervised algorithms. Nevertheless, as mentioned above, our complete algorithm surpasses these results and presents improved performance on 12 out of 13 benchmarks.
This analysis confirms our claim that full utilization of unlabeled data requires a semi-supervised training scheme that pays attention to the multimodal structure.

At this point, the reader might wonder about the applicability of the current synthetic datasets. To answer this question, we examine an additional training configuration,
% In the first setting, we only have synthetic data and unlabeled real-world data (Synth + Real-U). These results show that real-world data can boost performance even if only the synthetic datasets have transcription. 
in which we utilize all labeled real-world and synthetic data along with the unlabeled real-world data (Real-L + Synth + Real-U). Harnessing also the synthetic data leads to average accuracy improvements of +0.9\% on common benchmarks and of +0.4\% on non-common.
These results indicate that the labeled real-world datasets of scene text recognition are still in the low data regime and, therefore, can be contributed by synthetic alternatives.

% \subsection{Handwriting Text Recognition}
% In the handwritten text evaluation, we follow the path of \cite{seqclr_aberdam2021sequence} and experiment with various ratios of labeled and unlabeled data. We examine the results of state-of-the-art supervised methods, as well as semi-supervised ones that also leverage the unlabeled portion. We consider the same semi-supervised approaches as in Scene Text. In this setting, we leverage the same augmentation used in SeqCLR~\cite{seqclr_aberdam2021sequence} in our training scheme. As can be seen in \cref{tab:eval-htr-sota}, \AlgoName{} improves performance across all tested configuration and leads to a new state-of-the-art performance on both IAM and RIMES.

\section{Ablation Studies}
In this section, we examine the impact of the main components of \AlgoName{}. All the experiments here use labeled and unlabeled real-world data (Real-L + Real-U). While, for brevity, we report only the weighted average accuracies on the common and non-common benchmarks.

\subsection{Vision Model Pretraining}
\label{subsec:vision_pretraining_ablation}
\input{tables/vision_ablation}

\paragraph{\textbf{Two-stage versus unified training.}} Standard contrastive learning methods~\cite{chen2020simple,he2020momentum,grill2020bootstrap,caron2020unsupervised}, including SeqCLR~\cite{seqclr_aberdam2021sequence}, are based on the two training stages of contrastive-based pretraining and fully-supervised fine-tuning. In contrast to this scheme, we propose a unified training stage for the vision model pretraining, combining the contrastive learning objective with the supervised cross-entropy loss, as defined in \cref{eq:vision_pretraining_loss}. By doing so, we maintain that the overall training algorithm consists of three stages, similar to the origin ABINet training~\cite{abinet_fang2021read}. 
We present in \cref{tab:vision_ablation} the experimental comparison between these training approaches. As shown, the unified training stage even leads to performance improvements.

\paragraph{\textbf{Augmentation.}} While handwritten text images, which SeqCLR~\cite{seqclr_aberdam2021sequence} mainly focuses on, are usually of light-color solid background, text-in-the-wild images are often colorful with unique backgrounds.
This substantial domain difference leads us to suggest stronger color-texture augmentations, which are needed for learning effective representations that focus on the text content and not the general visual attributes.
\cref{tab:vision_ablation} shows the performance gain of the proposed augmentation technique compared to the origin SeqCLR augmentation pipeline~\cite{seqclr_aberdam2021sequence}.

\subsection{Consistency Regularization}
\label{subsec:fusion_pretraining_ablation}

\paragraph{\textbf{Sequential consistency regularization loss.}}
\input{tables/consistency_loss_ablation}
Here, we examine the contributions of the different ingredients of the proposed sequential consistency loss, defined in \cref{eq:sequential_consistency}. In particular, we compare cross-entropy with KL-divergence and soft labels (probabilities vectors) versus one-hot labels. In addition, we investigate stopping the teacher gradients and using a threshold.
In these experiments, the vision model is the teacher modality that generates the pseudo-labels, and all the modalities are the students trained on these pseudo-labels.
As shown in \cref{tab:consistency_loss_ablation}, in our setting, KL divergence as the core loss function instead of cross-entropy does not lead to consistent improvements.
Moreover, adding a certainty-based threshold for the teacher predictions (set to $90\%$) stabilizes results; however, it has a marginal effect in the best configuration that consists of cross-entropy, stop gradients, and uses one-hot (hard) labels.

\paragraph{\textbf{Teacher-student modalities.}} 
\input{tables/teacher_student_ablation}
In unimodal architectures, the teacher and student identity are usually clear since such models contain a single output. Nevertheless, in multimodal architectures, each modality output can be the teacher or the student. In \cref{tab:teacher_student}, we examine several of the possible teacher-student configurations and reveal that the most effective learning occurs when each modality is a teacher of its own. Meaning, that each modality generates its own pseudo-labels to train on, as defined in \cref{eq:overall_consistency_loss}.

\section{Conclusions and Future Work}

We present \AlgoName{}, the first multimodal tailored, semi-supervised learning algorithm for text recognition. To this end, we introduce a novel contrastive-based visual representation learning for scene text and a sequential, character-level consistency regularization. Although it leverages unlabeled data in addition to labeled one, our method maintains a compact three-stage algorithm. Comprehensive experiments demonstrate that \AlgoName{} outperforms supervised and semi-supervised state-of-the-art methods on multiple scene text recognition benchmarks. We believe that the success of \AlgoName{} will encourage researchers to further explore schemes for leveraging labeled and unlabeled real-world data for scene text recognition. In addition, our work sheds light on the status of current synthetic and real-world datasets, which can help future dataset creators. 

\clearpage
\bibliographystyle{splncs04}
\bibliography{egbib}

\clearpage

\appendix

\input{appendix_content}

\end{document}

%% file: math_commands.tex
\usepackage{amsmath,amsfonts,bm}

\def\eqref#1{equation~\ref{#1}}
\def\1{\bm{1}}

\newcommand{\test}{\mathcal{D_{\mathrm{test}}}}

\def\rvu{{\mathbf{i}}}

\def\rvu{{\mathbf{u}}}
\def\rvv{{\mathbf{v}}}

\def\rvy{{\mathbf{y}}}
\def\rvz{{\mathbf{z}}}

\def\rmF{{\mathbf{F}}}
\def\rmG{{\mathbf{G}}}

\def\rmK{{\mathbf{K}}}

\def\rmQ{{\mathbf{Q}}}

\def\rmW{{\mathbf{W}}}
\def\rmX{{\mathbf{X}}}
\def\rmY{{\mathbf{Y}}}

\DeclareMathAlphabet{\mathsfit}{\encodingdefault}{\sfdefault}{m}{sl}
\SetMathAlphabet{\mathsfit}{bold}{\encodingdefault}{\sfdefault}{bx}{n}
\newcommand{\tens}[1]{\bm{\mathsfit{#1}}}

\def\tP{{\tens{P}}}

\def\gD{{\mathcal{D}}}

\def\gL{{\mathcal{L}}}

\def\gZ{{\mathcal{Z}}}

\newcommand{\R}{\mathbb{R}}

\newcommand{\softmax}{\mathrm{softmax}}
\newcommand{\sigmoid}{\sigma}

\newcommand{\norm}[1]{\left\lVert#1\right\rVert}

\DeclareMathOperator{\simop}{sim}

\def\U{{\mathcal{U}}}

%% file: tables/dataset_info_sidebyside.tex
% \begin{table}
    % \centering
    % \begin{adjustbox}{margin=0pt \smallskipamount,center}
    \makeatletter\def\@captype{table}\makeatother% "Change float to table"
    \caption{\textbf{Training datasets}}
    \label{tab:dataset_info}
        \begin{tabular}{@{\extracolsep{1pt}}lllc@{}}
        \toprule
        && \textbf{Dataset} & \textbf{\#Words} \\
        \midrule
        \parbox[t]{2mm}{\multirow{13}{*}{\rotatebox[origin=c]{90}{\textbf{Real-L}}}}
        && SVT~\cite{Wang2011bottom} & 231 \\
        && IIIT~\cite{Mishra2012sj} & 1794 \\
        && IC13~\cite{Karatzas2013ic13} & 763 \\
        && IC15~\cite{Karatzas2015ic15} & 3,710 \\
        && COCO-Text~\cite{veit2016coco} & 39K \\
        && RCTW~\cite{shi2017icdar2017} & 8,186 \\
        && Uber-Text~\cite{zhang2017uber} & 92K \\
        && ArT~\cite{chng2019icdar2019} & 29K \\
        && LSVT~\cite{sun2019icdar} & 34K \\
        && MLT19~\cite{nayef2019icdar2019} & 46K \\
        && ReCTS~\cite{zhang2019icdar} & 23K \\
        \cmidrule(l){2-4}
        && Total & 276K \\
        \midrule
        \parbox[t]{2mm}{\multirow{4}{*}{\rotatebox[origin=c]{90}{\textbf{Real-U}}}}
        && Book32~\cite{iwana2016judging} & 3.7M \\
        && TextVQA~\cite{singh2019towards} & 463K \\
        && ST-VQA~\cite{biten2019scene} & 69K \\
        \cmidrule(l){2-4}
        && Total & 4.2M \\
        \midrule
        \parbox[t]{2mm}{\multirow{3}{*}{\rotatebox[origin=c]{90}{\textbf{Synth}}}}
        && ST~\cite{gupta2016synthetic} & 7M \\
        && MJ~\cite{jaderberg2014synthetic} & 9M \\
        \cmidrule(l){2-4}
        && Total & 16M \\
        % \midrule
        % \parbox[t]{2mm}{\multirow{2}{*}{\rotatebox[origin=c]{90}{\textbf{HW}}}}
        % && IAM (Eng)~\cite{marti2002iam} & 80K \\
        % && RIMES (Fr)~\cite{grosicki2009icdar} & 52K \\
        \bottomrule
    \end{tabular}
    % \end{adjustbox}
    
% \end{table}

%% file: tables/sota_all_scene_text_datasets_v2.tex
\begin{table*}[t] 
% \captionsetup{font=tiny}
  \tabcolsep=0.13cm
    \begin{center}
    \caption{\textbf{Scene text SOTA comparison.} Scene text recognition accuracies (\%) over common and non-common public benchmarks. We show the number of words in each dataset below its title and present weighted (by size) average results on each set of datasets. % No lexicon is used. 
    The best performing result at each column is marked in bold. ``*'' refers to reproduced results and ``$^{\text{Git}}$'' to GitHub model
    }
        \begin{adjustbox}{width=1\linewidth}
        \begin{tabular}{@{}l|cc|cccccc|c|ccccccc|c@{}}
            \toprule
            \multirow{3}{*}{\textbf{Method}} & \multirow{3}{*}{\textbf{Labeled}} & \multirow{3}{*}{\textbf{Unlabeled}} & \multicolumn{7}{c|}{\textbf{Common Benchmarks}} & \multicolumn{8}{c}{\textbf{Non-Common Benchmarks}} \\
            & \multirow{3}{*}{\textbf{Data}} & \multirow{3}{*}{\textbf{Data}} & \textbf{IIIT} & \textbf{SVT} & \textbf{IC13} & \textbf{IC15} & \textbf{SVTP} & \textbf{CUTE} & \textbf{Avg.} & \textbf{COCO} & \textbf{RCTW} & \textbf{Uber} & \textbf{ArT} & \textbf{LSVT} & \textbf{MLT19} & \textbf{ReCTS} & \textbf{Avg.}\\
            \ & \ & \ & 3,000 & 647 & 1,015 & 2,077 & 645 & 288 & 7,672 & 9,835 & 1,050 & 80,826 & 35,284 & 4,257 & 5,693 & 2,592 & 139,537 \\
            \midrule
            PlugNet \cite{litman2020scatter} & Synth & \ding{55} & 94.4 & 92.3 & 95.0 & 82.2 & 84.3 & 85.0 & 89.8 & - & - & - & - & - & - & - & -\\
            RobustScanner \cite{yue2020robustscanner} & Synth & \ding{55} & 95.3 & 88.1 & 94.8 & 77.1 & 79.5 & 90.3 & 88.2 & - & - & - & - & - & - & - & -\\
            SCATTER \cite{litman2020scatter} & Synth & \ding{55} & 93.7 & 92.7 & 93.9 & 82.2 & 86.9 & 87.5 & 89.7 & - & - & - & - & - & - & - & -\\
            Plugnet \cite{mou2020plugnet} & Synth & \ding{55} & 94.4 & 92.3 & 95.0 & 82.2 & 84.3 & 85.0 & 89.8 & - & - & - & - & - & - & - & -\\
            SRN \cite{srn_yu2020towards} & Synth & \ding{55} & 94.8 & 91.5 & 95.5 & 82.7 & 85.1 & 87.8 & 90.3 & - & - & - & - & - & - & - & -\\
            % Bhunia et al.~\cite{bhunia2021text} & Synth & \ding{55} & 92.3 & 89.9 & 93.3 & 76.9 & 84.4 & 86.3 & 87.2 & - & - & - & - & - & - & - & -\\
            VisionLAN \cite{wang2021two} & Synth & \ding{55} & 95.8 & 91.7 & 95.7 & 83.7 & 86.0 & 88.5 & 91.1 & - & - & - & - & - & - & - & -\\
            \midrule
            TRBA \cite{Baek2019clova} & Synth & \ding{55} & 92.1 & 88.9 & 93.1 & 74.7 & 79.5 & 78.2 & 85.7 & 50.2 & 59.1 & 36.7 & 57.6 & 58.0 & 80.3 & 80.6 & 46.3\\
            TRBA \cite{whatif_baek2021if} & Real-L & \ding{55} & 93.5 & 87.5 & 92.6 & 76.0 & 78.7 & 86.1 & 86.6 & 62.7 & 67.7 & 52.7 & 63.2 & 68.7 & 85.8 & 83.4 & 58.6\\
            TRBA$_{PL}$ \cite{whatif_baek2021if} & Real-L & Real-U & 94.8 & 91.3 & 94.0 & 80.6 & 82.7 & 88.1 & 89.3 & 66.9 & 71.5 & 54.2 & 66.7 & 73.5 & 87.8 & 85.6 & 60.9\\
            TRBA$_{PL}$ \cite{whatif_baek2021if} & Real-L,Synth & Real-U & 95.2 & 92.0 & 94.7 & 81.2 & 84.6 & 88.7 & 90.0 & - & - & - & - & - & - & - & -\\
            TRBA$_{PL}$ \cite{whatif_baek2021if} & Real-L,Synth & Real-U & 95.2 & 92.0 & 94.7 & 81.2 & 84.6 & 88.7 & 90.0 & - & - & - & - & - & - & - & -\\
            \midrule
            ABINet$^{\text{Git}}$ \cite{abinet_fang2021read} & Synth & \ding{55} & 96.4 & 93.2 & 95.1 & 82.1 & 89.0 & 89.2 & 91.2 & 63.1 & 59.7 & 39.6 & 68.3 & 59.5 & 85.0 & 86.7 & 52.0 \\
            ABINet* \cite{abinet_fang2021read} & Real-L & \ding{55} & 95.5 & 93.4 & 94.4 & 83.0 & 87.1 & 89.6 & 90.8 & 69.2 & 71.6 & 55.7 & 67.7 & 73.7 & 88.2 & 90.6 & 62.4 \\
            % ABINet* \cite{abinet_fang2021read} (bug) & Real-L & Real-U & 96.5 & 94.6 & 94.7 & 83.6 & 88.1 & 91.7 & 91.7 & 71.3 & 73.5 & 56.8 & 69.5 & 74.8 & 89.1 & 91.9 & 63.7\\
            ABINet$_{PL}$* \cite{abinet_fang2021read,whatif_baek2021if} & Real-L & Real-U & 96.4 & 94.1 & 95.0 & 83.7 & 88.8 & 93.1 & 91.8 & 71.2 & 74.2 & 56.8 & 70.5 & 75.0 & 89.1 & 90.9 & 63.9 \\
            ABINet$_{est}$* \cite{abinet_fang2021read} & Real-L & Real-U & 96.5 & \textbf{96.3} & 95.7 & 83.7 & 89.1 & 92.0 & 92.1 & 71.7 & 73.8 & 56.8 & 70.1 & 75.7 & 89.3 & 91.6 & 63.9 \\
            \midrule
            % Ours & MJ+ST & - & ? & ? & ? & ? & ? & ? & ? & ? & ? & ? & ? & ? & ? & ? & ?\\
            % Ours & Real-L & - & ? & ? & ? & ? & ? & ? & ? & ? & ? & ? & ? & ? & ? & ? & ?\\
            % Ours only Vision (old) & Real-L & Real-U & 96.3 & 94.9 & 96.2 & 84.4 & 89.8 & 93.4 & 92.3 & 71.4 & 74.7 & 57.3 & 69.7 & 75.8 & 89.6 & 92.7 & 64.1 \\
            SemiMTR-V & Real-L & Real-U & 95.6 & 93.5 & 95.2 & 82.5 & 88.1 & 90.6 & 91.0 & 70.5 & 75.1 & 57.7 & 69.5 & 75.2 & 89.6 & 92.3 & 64.2 \\
            SemiMTR-F & Real-L & Real-U & 96.5 & 95.4 & 96.5 & 84.2 & 89.6 & 90.6 & 92.3 & 70.9 & 74.9 & 57.7 & 70.3 & 75.5 & 89.3 & 91.5 & 64.4 \\
            % Ours & Real-L & Real-U & \textbf{96.8} & \textbf{95.8} & \textbf{96.4} & \textbf{84.2} & \textbf{89.5} & \textbf{91.7} & \textbf{92.4} & 71.2 & \textbf{74.4} & \textbf{58.0} & \textbf{70.5} & \textbf{75.7} & \textbf{89.9} & \textbf{92.1} & \textbf{64.7}\\
            % Ours (old) & Real-L & Real-U & 96.7 & 95.8 & 96.3 & 83.7 & 89.1 & 91.7 & 92.2 & 71.3 & 74.1 & 58.0 & 70.5 & 75.7 & 89.9 & 91.9 & 64.7 \\
            % Ours & Real-L & Real-U & 96.2 & 96.1 & 96.3 & 83.8 & 88.7 & 94.1 & 92.1 & 71.4 & 75.1 & 58.5 & 70.5 & 76.9 & 89.9 & 92.4 & 65.0 \\
            % Ours & Real-L & Real-U & 96.2 & 95.8 & 96.4 & 84.2 & 88.7 & 92.7 & 92.2 & 71.0 & 75.5 & 58.5 & 70.5 & 76.8 & 90.0 & 92.4 & 65.0 \\
            % Ours & Real-L & Real-U & 96.3 & \textbf{95.7} & 96.3 & 83.9 & \textbf{90.4} & \textbf{93.8} & \textbf{92.3} & \textbf{71.7} & \textbf{75.3} & \textbf{58.3} & \textbf{71.0} & \textbf{76.7} & \textbf{89.6} & \textbf{92.1} & \textbf{65.1} \\
            % Ours (all2all) & Real-L & Real-U & 96.4 & 95.8 & 96.4 & 84.4 & 89.8 & 92.4 & 92.4 & 71.5 & 74.9 & 58.6 & 70.5 & 76.8 & 90.0 & 92.4 & 65.1 \\
            SemiMTR & Real-L & Real-U & \textbf{96.7} & 95.5 & \textbf{96.6} & \textbf{83.8} & \textbf{90.5} & \textbf{93.8} & \textbf{92.4} & \textbf{72.0} & \textbf{75.8} & \textbf{58.5} & \textbf{70.8} & \textbf{77.1} & \textbf{90.3} & \textbf{92.5} & \textbf{65.2} \\
            \cdashline{1-18}
            % SemiMTR & Synth & Real-U & 93.4 & 96.9 & 96.5 & 79.9 & 90.5 & 89.2 & 90.0 & 63.1 & 64.5 & 45.3 & 65.0 & 67.7 & 83.9 & 88.8 & 54.7 \\
            SemiMTR & Real-L,Synth & Real-U & \textbf{97.3} & \textbf{96.6} & \textbf{97.0} & \textbf{84.7} & \textbf{93.0} & \textbf{93.8} & \textbf{93.3} & \textbf{72.7} & \textbf{76.3} & 58.4 & \textbf{72.3} & \textbf{77.1} & 90.2 & \textbf{93.2} & \textbf{65.6} \\
            \bottomrule
        \end{tabular}
        \end{adjustbox}
    \vspace{+1mm}

    \label{tab:eval-str-sota}
    \end{center}
\end{table*}

%% file: tables/vision_ablation.tex
\begin{table*}[t] 
% \captionsetup{font=tiny}
  \tabcolsep=0.13cm
    \begin{center}
    \caption{\textbf{Ablation of vision model pretraining.} Accuracy of the vision model when pretrained via a typical two-stage scheme versus our unified method and when using different augmentation pipelines
    }
        \begin{adjustbox}{width=0.7\linewidth}
        \begin{tabular}{@{}cccc@{}}
            \toprule
            \multirow{3}{*}{\textbf{Method}} & \multirow{3}{*}{\textbf{Augmentations}} & \multicolumn{2}{c}{\textbf{Vision Model}} \\
            & &  Common & Non-Common \\
            & &  Benchmarks & Benchmarks \\
            \midrule
            \textit{Supervised Baseline}~\cite{abinet_fang2021read} & ABINet~\cite{abinet_fang2021read} & 85.3 & 57.9 \\
            % Two-Stage & SeqCLR~\cite{seqclr_aberdam2021sequence} & ? & ? \\
            Two-Stage & SeqCLR~\cite{seqclr_aberdam2021sequence} & 86.7 & 59.6 \\
            Unified & SeqCLR~\cite{seqclr_aberdam2021sequence} & 87.0 & 60.2 \\
            Unified & Ours & \textbf{88.1} & \textbf{60.3} \\
            \bottomrule
        \end{tabular}
        \end{adjustbox}
    % \vspace{+1mm}
    \label{tab:vision_ablation}
    \end{center}
\end{table*}

%% file: tables/consistency_loss_ablation.tex
\begin{table*}[t] 
% \captionsetup{font=tiny}
  \tabcolsep=0.13cm
    \begin{center}
    \caption{\textbf{Consistency loss ablation.} Accuracy of different consistency loss configurations, calculated on both the common and non-common benchmarks
    }
        \begin{adjustbox}{width=0.7\linewidth}
        \begin{tabular}{@{}cccccc@{}}
            \toprule
            \textbf{Consistency} & \textbf{Stop} & \textbf{Soft} & \multirow{2}{*}{\textbf{Threshold}} & \textbf{Common} & \textbf{Non-Common} \\
            \textbf{Loss} & \textbf{Gradients} & \textbf{labels} &  & \textbf{Benchmarks} & \textbf{Benchmarks} \\
            \midrule
            CE & \cmark  &  \xmark  &  \xmark  & 91.8 & \textbf{65.0}\\
            CE  &  \xmark  &  \cmark  &  \xmark  & 91.8 & 64.1 \\
            % CE  &  \cmark  &  \xmark  &  \xmark  & 92.0 & 65.0 \\
            CE  &  \cmark  &  \cmark  &  \xmark  & 91.8 & 64.5 \\
            KL Divergence  &  \xmark  &  \cmark  &  \xmark  & \textbf{92.2} & 64.5 \\
            \midrule
            CE  &  \cmark  &  \xmark  &  \cmark  & \textbf{92.2} & \textbf{65.0} \\
            CE  &  \xmark  &  \cmark  &  \cmark  & 92.0 & \textbf{65.0} \\
            % CE  &  \cmark  &  \xmark  &  \cmark  & 92.1 & \textbf{65.0} \\
            CE  &  \cmark  &  \cmark  &  \cmark  & 92.0 & 64.7 \\
            KL Divergence  &  \xmark  &  \cmark  &  \cmark  & 92.1 & 64.8 \\
            \bottomrule
        \end{tabular}
        \end{adjustbox}
    % \vspace{+1mm}
    \label{tab:consistency_loss_ablation}
    \end{center}
\end{table*} 

%% file: tables/teacher_student_ablation.tex
% \begin{table*}[t] 
% % \captionsetup{font=tiny}
%   \tabcolsep=0.13cm
%     \begin{center}
%         \begin{adjustbox}{width=1.0\linewidth}
%         \begin{tabular}{@{}ll|ccc|c@{}}
%             \toprule
%             Teacher & Student/s & Regular text & Irregular text & Other & Avg.\\
%             \midrule
%             Fusion & Vision & 96.0 & 85.9 & 64.7 & 66.1 \\
%             Fusion & Language & 95.9 & 85.2 & 64.6 & 66.0 \\
%             Fusion & Fusion & 96.1 & 86.2 & 65.1 & 66.5 \\
%             Fusion & Vision, Language, Fusion & 96.1 & 86.4 & 65.1 & 66.5 \\
%             \midrule
%             Vision & Vision & 95.9 & 85.3 & 64.8 & 66.2 \\
%             Vision & Language & ? & ? & ? & ?\\
%             Vision & Fusion & 96.3 & 85.5 & 65.2 & 66.6 \\
%             Vision & Vision, Language, Fusion & 96.0 & 85.3 & 65.0 & 66.4 \\
%             \midrule
%             Vision, Language, Fusion & Vision, Language, Fusion & 96.3 & 86.3 & 65.1 & 66.5 \\
%             \bottomrule
%         \end{tabular}
%         \end{adjustbox}
%     % \vspace{+1mm}
%     \caption{Accuracy of seven evaluation datasets: COCO, RCTW, Uber, ArT, LSVT, MLT19, and ReCTS.
%     Avg. denotes the averaged accuracies of seven datasets.
%     The number of evaluation sets of each datasets is described in Table~\ref{sup-tab:split}.
%     }
%     \label{sup-tab:eval-seven-addition}
%     \end{center}
% \end{table*}

\begin{table*}[t] 
% \begin{wraptable}{htbr}{0.5\linewidth}
% \captionsetup{font=tiny}
  \tabcolsep=0.13cm
    \begin{center}
    \caption{\textbf{Teacher-student identities.} Accuracy attained by varying teacher and student identities
    }
        \begin{adjustbox}{width=0.8\linewidth}
        \begin{tabular}{@{}cccc@{}}
            \toprule
            
            \multirow{2}{*}{\textbf{Teacher}} & \multirow{2}{*}{\textbf{Student}}
            & \textbf{Common} & \textbf{Non-Common} \\
            & & \textbf{Benchmarks} & \textbf{Benchmarks}  \\
            \midrule
            Vision  &  Vision  & 91.7 & 64.8 \\
            Vision & Language & 91.9 & 64.5 \\
            Vision  &  Fusion  & 92.1 & \textbf{65.2}  \\
            Vision  &  Vision, Language, Fusion  & 91.8 & 65.0  \\
            \midrule
            Fusion  &  Vision  & 92.0 & 64.7 \\
            Fusion  &  Language  & 91.7 & 64.6 \\
            Fusion  &  Fusion  & 92.2 & 65.1 \\
            Fusion  &  Vision, Language, Fusion  & 92.3 & 65.1 \\
            \midrule
            Vision, Language, Fusion  &  Vision, Language, Fusion  & \textbf{92.4} & \textbf{65.2} \\
            \bottomrule
        \end{tabular}
        \end{adjustbox}
    % \vspace{+1mm}
    \label{tab:teacher_student}
    \end{center}
% \end{wraptable}
\end{table*}

% \begin{table*}[t] 
% % \begin{wraptable}{htbr}{0.5\linewidth}
% % \captionsetup{font=tiny}
%   \tabcolsep=0.13cm
%     \begin{center}
%         \begin{adjustbox}{width=1\linewidth}
%         \begin{tabular}{@{}ll|cc|c@{}}
%             \toprule
            
%             \multirow{2}{*}{\textbf{Teacher}} & \multirow{2}{*}{\textbf{Student}}
%             & Common & Non-Common & \multirow{2}{*}{Avg.} \\
%             & & Benchmarks & Benchmarks  \\
%             \midrule
%             Fusion  &  Vision  & 92.0 & 64.7 & 66.1 \\
%             Fusion  &  Language  & 91.7 & 64.6 & 66.0 \\
%             Fusion  &  Fusion  & 92.2 & 65.1 & 66.5 \\
%             Fusion  &  Vision, Language, Fusion  & 92.3 & 65.1 & 66.5 \\
%             \midrule
%             Vision  &  Vision  & 91.7 & 64.8 & 66.2 \\
%             Vision & Language & ? & ? & ?\\
%             Vision  &  Fusion  & 92.1 & \textbf{65.2} & \textbf{66.6} \\
%             Vision  &  Vision, Language, Fusion  & 91.8 & 65.0 & 66.4 \\
%             \midrule
%             Vision, Language, Fusion  &  Vision, Language, Fusion  & \textbf{92.4} & \textbf{65.2} & \textbf{66.6} \\
%             \bottomrule
%         \end{tabular}
%         \end{adjustbox}
%     % \vspace{+1mm}
%     \caption{Accuracy of seven evaluation datasets: COCO, RCTW, Uber, ArT, LSVT, MLT19, and ReCTS.
%     Avg. denotes the averaged accuracies of seven datasets.
%     The number of evaluation sets of each datasets is described in Table~\ref{sup-tab:split}.}
%     \label{sup-tab:eval-seven-addition}
%     \end{center}
% % \end{wraptable}
% \end{table*}

%% file: appendix_content.tex
\section{ABINet Architecture}
\label{app:abinet}

\begin{figure}
\normalsize
  \centering
  \includegraphics[width=0.95\textwidth]{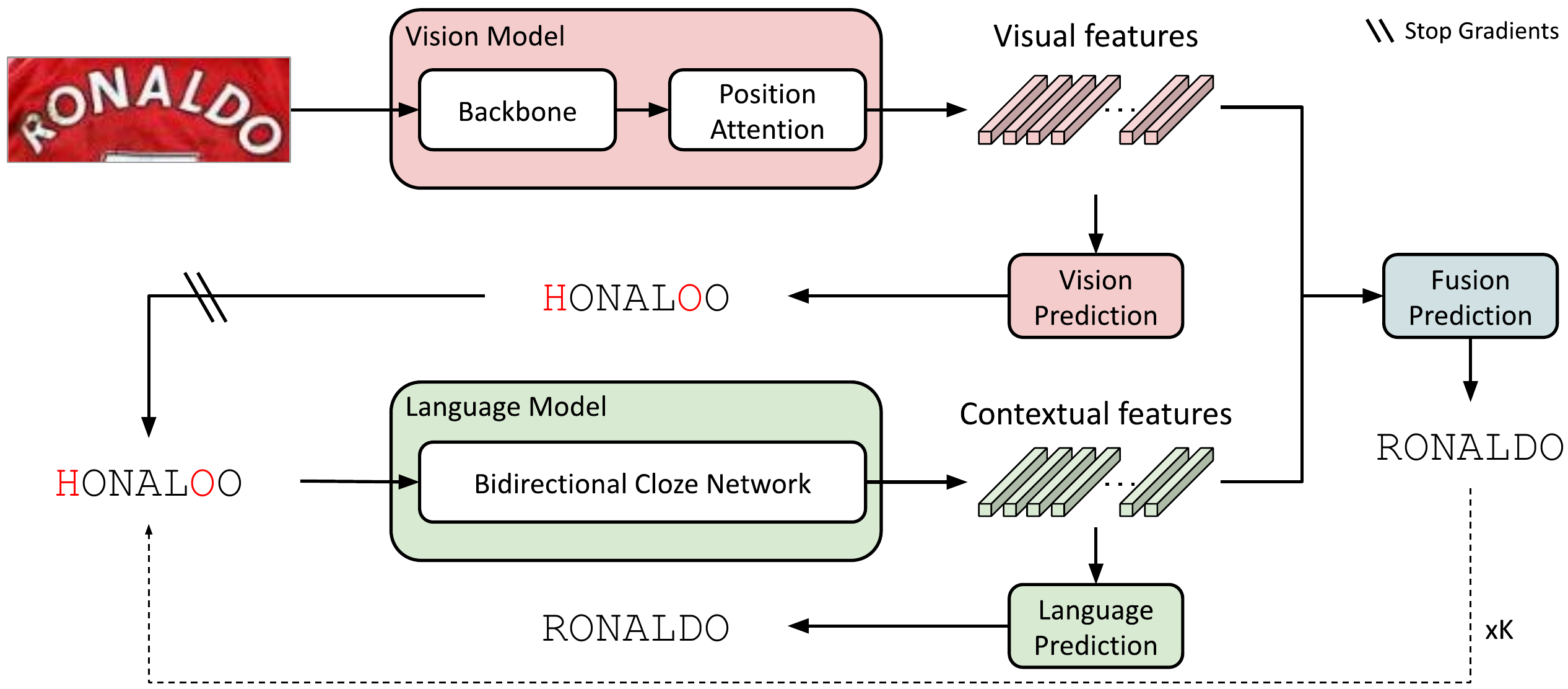}
  \caption{\textbf{ABINet text recognition architecture}. As mentioned in the main paper and repeated here for convenient, we adopt ABINet~\cite{abinet_fang2021read} as our case study of vision-language multimodal recognizers.
  In this scheme, the vision model first extracts a sequence of visual features from a given image and then decodes it into character predictions. Next, these predictions are fed to the language model, which derives contextual features. Finally, the fusion model operates on the visual and contextual features and provides the network final prediction. To further improve accuracy, an optional iterative phase, drawn with a dashed line, reinserts the output back to the language model as a predictions-refinement step}
  \label{fig:abinet_background_app}
\end{figure}

In our work we adopt the ABINet-LV, which depicted here again, \cref{fig:abinet_background_app}, for convenient. This architecture consists of the following three models:
\begin{enumerate}
    \item \textbf{Vision Model} consists of backbone network and position attention module. The visual backbone comprises ResNet~\cite{ASTER} and Transformer units~\cite{lyu20192d,srn_yu2020towards}. Thus, the visual backbone features for a given image $\rmX \in \R^{3 \times H \times W}$ are:
    \begin{equation}
        \rmF_b = \operatorname{Transformer}(\operatorname{ResNet}(\rmX)) \in \R^{\frac{H}{4} \times \frac{W}{4} \times C},
    \end{equation}
    where $C$ is the feature dimension.
    Then, the position attention module, which is based on the query paradigm~\cite{vaswani2017attention}, is applied on $\rmF_b$ to produce the visual features:
    \begin{equation}
        \rmF_v = \softmax\left( \frac{\rmQ \rmK^T}{\sqrt{C}} \right) \operatorname{Flatten}(\rmF_b) \in \R^{T \times C},
    \end{equation}
    where $\rmK = \operatorname{Mini\,U-Net}(\rmF_b) \in \R^{\frac{HW}{16} \times C}$~\cite{ronneberger2015u}, $T$ is the character sequence length, and $\rmQ \in \R^{T \times C}$ is the positional encoding of the character orders.
    Finally, a sequence of linear classifier and softmax operator transcribes the visual features into character probabilities in parallel.
    These character probabilities are fed to the language model as a seed text generator.
    
    % to embed an image and provide an initial sequence of characters. In particular, given an image $\rmX \in \R^{C \times H \times W_i}$, where $C$ denotes the number of input channels, $H$ the image height, and $W_i$ the width of each image which may vary. ResNet~\cite{ASTER} and Transformer units~\cite{lyu20192d,srn_yu2020towards} are employed as the feature extraction network, producing $\rmV_B \in \R^{F \times \frac{H}{4} \times \frac{W}{4}}$. Next, the position attention module is employed, yielding $\rmV_V$. The position attention module is based on the query paradigm~\cite{vaswani2017attention} and leverages a mini U-Net~\cite{ronneberger2015u}.
    % A linear layer is used to transcribe $\rmV_V$ in parallel into a seed character probabilities of length $T$, namely $\rmY_V$. A supervised cross-entropy loss is employed at this point.
    \item \textbf{Language Model} is a variant of $L$-layers transformer decoder ($L = 4$ in our case), named bidirectional cloze network (BCN). The input to this model is the visual predictions in the first iteration, while in next iterations is the fusion model predictions.
    As before, a linear layer and softmax operates on the contextual features, $\rmF_l \in \R^{T \times C}$ to provide character probabilities. We refer the reader to \cref{app:language_pretraining} for additional details. 
    % uses the seed text $\rmY_V$ as input and outputs a bidirectional contextualized representations $\rmV_L^i, i \in K$ where $K$ is the number of iterative correction. Again, a linear layer is employed to transcribe $\rmV_L^i$ into a language prediction of length $T$, namely $\rmY_L^i$. As the input for this stage is just a sequence of character probabilities, this stage is autonomous and can be trained using unlabeled raw text. This module is trained using a variation of the masked language model scheme proposed by \cite{abinet_fang2021read} with a supervised cross-entropy loss.
    
    \item \textbf{Fusion Model}, a gated mechanism which operates on the visual and contextual features and produces the align features.
    As depicted in \cref{fig:abinet_background_app}, the fusion predictions can be inserted again to the language model in an iterative correction manner. Therefore, this model can be written as,
    \begin{eqnarray}
        \rmG & = & \sigmoid \left( \left[ \rmF_v, \rmF_l^i \right]  \rmW_f \right) \\
        \rmF_f^i & = & \rmG \odot \rmF_v + (1 - \rmG) \odot \rmF_l^i,
    \end{eqnarray}
    where $\odot$ denotes the Hadamard (element-wise) product and $i$ is the iteration number out of $K$ total ($K = 3$ in our case). 
    % to produce $\rmY_F^i$. The last two stage are performed $K$ times in an iterative manner, in which the $i$th language model takes as input $\rmY_F^{i-1}$. Similarly to the other stages, a supervised cross-entropy loss is used as supervision.   
\end{enumerate}
We refer the reader to ABINet work~\cite{abinet_fang2021read} for more details.

\section{Language Model Pretraining}
\label{app:language_pretraining}
One of the main themes advocated in ABINet~\cite{abinet_fang2021read} is that the language model (LM) should be decoupled from the vision one and trained explicitly to learn linguistic rules. Such modality-separation enables to harness the remarkable improvements attained in training LM in Natural Language Processing and to pretrain it on a large text corpus. More specifically, the LM pretraining is based on bidirectional representation learning using a masked language model (MLM). A possible, straightforward way to do so is by masking a certain input text token and training the LM to predict it. Nevertheless, doing so requires applying such masking $n$ for a sentence of length $n$. To mitigate this overhead, ABINet proposes to apply the masking on the attention maps of the LM architecture -- when predicting output $i$, the attention mask prevents the model from receiving information from input $i$, and thus, implements the MLM efficiently.

The LM is trained using MLM on a filtered version of WikiText-103~\cite{merity2016pointer} corpus (applying basic filtering -- word's size and unsupported characters), that contains 86.5 million text tokens. The model is pretrained for 80 epochs, using a batch size of 4096, Adam optimizer ($\beta_1=0.9, \beta_2=0.999$), no weight decay, gradient clipping of 20, and initial learning rate of 0.0001 with a drop (multiplying it by $0.1$) at epoch 70.

\section{Comparison to State-of-the-art Semi-Supervised}
\label{app:comparison_sota}
\textbf{Scene Text}: In this section, we provide additional implementation and technical details, regarding the reproduction of the semi-supervised methods, denoted as ABINet$_{PL}$~\cite{whatif_baek2021if} and ABINet$_{est}$~\cite{abinet_fang2021read}. In both of these, we fine-tune ABINet using a pretrained language model and a pretrained visual model, trained on the labeled portion $(X^l,Y^l)$. Then, we use it to predict the unlabeled portion $X^u$ and construct a unified dataset, wherein ABINet$_{est}$, we apply certainty-based filtering, as proposed in \cite{abinet_fang2021read}. 
More specifically, we filter out pseudo labels with a text certainty $\mathcal{C}$ below predefined value Q (we set $Q=0.9$, as in the original paper). The text certainty of a pseudo label is defined as the minimal certainty of its characters, denoted as $c(t)$. The characters' certainty is the maximal over the prediction of all the iterations of ABINet (the certainty of the $i$-th character of the $k$-th iteration is denoted as $c(t,k)$). This filtering mechanism description is depicted in Equation \ref{eq:abinet_est} below.

\begin{equation}
    \label{eq:abinet_est}
        \begin{gathered}
            c(t)=\max_{1\leq k \leq K} c(t,k) \\
            \mathcal{C}=\min_{1\leq t \leq T} c(t)
        \end{gathered}
\end{equation}

\noindent Where K is the number of ABINet's iterations and T is the effective length of the pseudo label. Finally, we use the pretrained vision and language models and fine-tune the ABINet using the unified dataset. These semi-supervised methods are described in \cref{alg:abinet_pl,alg:abinet_est}.

\begin{minipage}{0.46\textwidth}
\begin{algorithm}[H]
    \caption{ABINet$_{PL}$}
    \label{alg:abinet_pl}
    \noindent \textbf{Inputs:} Labeled data
        ($X^{l}, Y^{l}$) and unlabeled data $X^{u}$.
    \begin{algorithmic}[1]
        \State Pretrain vision model on ($X^{l}, Y^{l}$)
        \State Fine-tune ABINet on ($X^{l}, Y^{l}$)
        \State Use ABINet to generate pseudo-labels ($Y^{u}$) on $X^{u}$
        \State $(X,Y) \leftarrow \{(X^{l}, Y^{l}) \cup (X^{u}, Y^{u})\}$
        \State Retrain ABINet on ($X,Y$)
    \end{algorithmic}
\end{algorithm}
\end{minipage}
\hfill
\begin{minipage}{0.46\textwidth}
\begin{algorithm}[H]
    \caption{ABINet$_{est}$}
    \label{alg:abinet_est}
    \noindent \textbf{Inputs:} Labeled data
        ($X^{l}, Y^{l}$) and unlabeled data $X^{u}$.
    \begin{algorithmic}[1]
        \State Pretrain vision model on ($X^{l}, Y^{l}$)
        \State Fine-tune ABINet on ($X^{l}, Y^{l}$)
        \State Use ABINet to generate pseudo-labels ($Y^{u}$) on $X^{u}$
        \State Filter $(X^{u}, Y^{u})$ using \cref{eq:abinet_est}
        \State $(X,Y) \leftarrow \{(X^{l}, Y^{l}) \cup (X^{u}, Y^{u})\}$
        \State Retrain ABINet on ($X,Y$)
    \end{algorithmic}
\end{algorithm}
\end{minipage}

\section{Data}
\label{app:dtatsets}
\begin{figure}[t]
\normalsize
  \centering
  \includegraphics[width=0.9\textwidth]{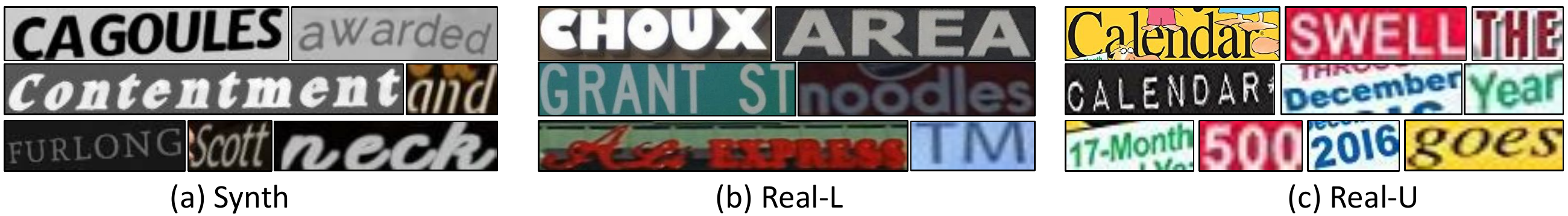}
  \caption{\textbf{Dataset samples.} We provide examples from each of the datasets used in this work, namely; (a) Synth, (b) Real-L and (c) Real-U}
  \label{fig:app_dataset_samples}
\end{figure}

As mentioned in the main paper, in this work we consider the same datasets used by Beak et al.~\cite{whatif_baek2021if}. In particular, we use the Synth, Real-L and Real-U datasets, see examples from each in \cref{fig:app_dataset_samples}. 

\section{Augmentations}
\label{app:augmentations}

% \begin{itemize}
%     \item Channel shuffle.
%     \item K-means color quantization.
%     \item Gray scale, gamma contrast and histogram equalization.
%     \item Blur, Gaussian noise and element-wise multiplication.
%     \item Hue, saturation and color temperature changes.
% \end{itemize}

A pseudo-code for the augmentation pipeline, written with the \texttt{imgaug}~\cite{imgaug} package, is as follows.
\begin{lstlisting}[language=Python]
from imgaug import augmenters as iaa

iaa.Sequential([
    iaa.Invert(0.5),
    iaa.OneOf([
        iaa.ChannelShuffle(0.35),
        iaa.Grayscale(alpha=(0.0, 1.0)),
        iaa.KMeansColorQuantization(),
        iaa.HistogramEqualization(),
        iaa.Dropout(p=(0, 0.2), per_channel=0.5),
        iaa.GammaContrast((0.5, 2.0)),
        iaa.MultiplyBrightness((0.5, 1.5)),
        iaa.AddToHueAndSaturation((-50, 50), per_channel=True),
        iaa.ChangeColorTemperature((1100, 10000))
    ]),
    iaa.OneOf([
        iaa.Sharpen(alpha=(0.0, 0.5), lightness=(0.0, 0.5)),
        iaa.OneOf([
            iaa.GaussianBlur((0.5, 1.5)),
            iaa.AverageBlur(k=(2, 6)),
            iaa.MedianBlur(k=(3, 7)),
            iaa.MotionBlur(k=5)
        ])
    ]),
    iaa.OneOf([
        iaa.Emboss(alpha=(0.0, 1.0), strength=(0.5, 1.5)),
        iaa.AdditiveGaussianNoise(scale=(0, 0.2 * 255)),
        iaa.ImpulseNoise(0.1),
        iaa.MultiplyElementwise((0.5, 1.5))
    ])
])
\end{lstlisting}

% \section{EMA}
% \label{app:ema_ablation}
% \input{tables/ema_ablation}

% \section{Fusion Augmentations Study}
% \label{app:fustion_augs_ablation}
% \input{tables/fusion_augs_ablation}

\section{Implementation Details}
\label{app:implementation_details}

As baseline, we use the code of ABINet\footnote{https://github.com/FangShancheng/ABINet}~\cite{abinet_fang2021read}, and our architectural changes are implemented on top of ABINet-LV.
All experiments are trained and tested using the PyTorch\footnote{https://pytorch.org/} framework on 4 Tesla V100 GPUs with 16GB memory each. The model dimension $C$ is set to 512 throughout. There are 4 layers in BCN with 8 attention heads each layer. Images are directly resized to 32 × 128. We train using the ADAM optimizer with the initial learning rate of $1e^{-4}$, decay rate of $0.1$, and gradient clipping of a magnitude of $20$. The number of character classes is 37; 10 for digits, 26 for alphabets, and a single padding token. All the coefficient scalars are set to 1.

For the vision pretraining, the batch size is 304 with a sampling ratio of 30\%, 70\% between the labeled and unlabeled data, respectively. The number of epochs is $25$ with scheduler periods of $[17, 5, 3]$. The SeqCLR head~\cite{seqclr_aberdam2021sequence} does not contain a projection head, the temperature parameter is $\tau = 0.1$, and the window-to-instance mapping function outputs $T = 5$ instances for each feature map.

We use the pretrained language model provide by \cite{abinet_fang2021read}. For the fusion training, the batch size is $232$ with a sampling ratio of 30\%, 70\% between the labeled and unlabeled data, respectively. The number of epochs is $5$ with scheduler periods of $[3, 1, 1]$. As for augmentation, we refer the reader to \cref{app:augmentations} for more details.